\documentclass{article} 
\usepackage{conference,times}

\usepackage{graphicx}
\usepackage{amsmath}
\usepackage{amssymb}

\usepackage{array}
\usepackage{booktabs}
\usepackage{multirow}
\usepackage{makecell}
\usepackage{ulem}
\usepackage[misc]{ifsym}

\usepackage{url}
\usepackage{multirow}
\usepackage{booktabs}
\usepackage{pifont}
\usepackage{amsmath}
\usepackage{algorithm}
\usepackage{mathtools}
\usepackage{algorithm}
\usepackage{algpseudocode}
\usepackage{listings}
\usepackage{wrapfig}
\usepackage{colortbl}
\usepackage{array}

\usepackage{rotating}
\usepackage{makecell}
\usepackage{color}

\usepackage{tabularx}
\newcolumntype{C}{>{\centering\arraybackslash}X}

\usepackage{amsmath,amsfonts,bm}

\def\eqref#1{equation~\ref{#1}}

\def\1{\bm{1}}

\DeclareMathAlphabet{\mathsfit}{\encodingdefault}{\sfdefault}{m}{sl}
\SetMathAlphabet{\mathsfit}{bold}{\encodingdefault}{\sfdefault}{bx}{n}

\usepackage{hyperref}
\usepackage{url}

\definecolor{top1}{RGB}{154,106,114}
\definecolor{top2}{RGB}{181,125,134}
\definecolor{top3}{RGB}{217,150,161}
\definecolor{top4}{RGB}{250,171,184}
\definecolor{top5}{RGB}{252,214,210}
\definecolor{top6}{RGB}{255,246,242}

\usepackage{xcolor,pifont}
\newcommand*\colourcheck[1]{%
	\expandafter\newcommand\csname #1check\endcsname{\textcolor{#1}{\ding{52}}}%
}
\colourcheck{blue}
\colourcheck{green}
\colourcheck{red}
\usepackage{makecell}

\newcommand{\modell}{\textit{MedSora~}}
\newcommand{\model}{\textit{MedSora}}

\usepackage[T1]{fontenc}

\usepackage{adjustbox}
\usepackage{multirow}
\usepackage{mathtools}
\usepackage{graphicx}
\usepackage[title]{appendix}%

\title{
	\begin{minipage}{0.08\textwidth}
		\includegraphics[width=\linewidth]{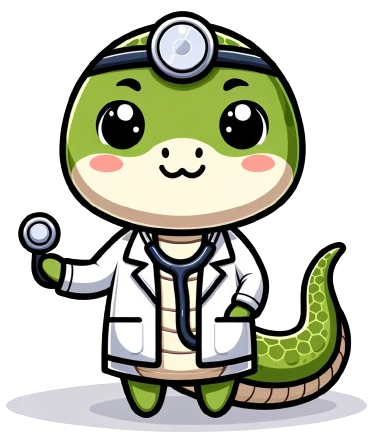}
	\end{minipage}
	\model: Optical Flow Representation Alignment \\ Mamba Diffusion Model for Medical Video Generation
}

\finalcopy

\author{Zhenbin Wang, Lei Zhang$^{(\textrm{\Letter})}$\thanks{$^{(\textrm{\Letter})}$The corresponding author.}, Lituan Wang, Minjuan Zhu, Zhenwei Zhang \\ 
Machine Intelligence Laboratory\\
College of Computer science, Sichuan University\\
Chengdu, Sichuan Province, China \\
\texttt{wangzhenbin@stu.scu.edu.cn, leizhang@scu.edu.cn} \\
}

\begin{document}

\maketitle

\begin{abstract}
Medical video generation models are expected to have a profound impact on the healthcare industry, including but not limited to medical education and training, surgical planning, and simulation.
Current video diffusion models typically build on image diffusion architecture by incorporating temporal operations (such as 3D convolution and temporal attention). Although this approach is effective, its oversimplification 
limits spatio-temporal performance and consumes substantial computational resources.
To counter this, we propose \textit{\textcolor{red}{Med}ical \textcolor{red}{S}imulation Vide\textcolor{red}{o} Gene\textcolor{red}{ra}tor} (\model), which incorporates three key elements:
\textit{i)} a video diffusion framework integrates the advantages of attention and Mamba, balancing low computational load with high-quality video generation,
\textit{ii)} an optical flow representation alignment method that implicitly enhances attention to inter-frame pixels,
and \textit{iii)} a video variational autoencoder (VAE) with frequency compensation addresses the information loss of medical features that occurs when transforming pixel space into latent features and then back to pixel frames.
Extensive experiments and applications demonstrate that \model~exhibits superior visual quality in generating medical videos, outperforming the most advanced baseline methods.
Further results and code are available at \url{https://wongzbb.github.io/MedSora/}.
\end{abstract}

\section{Introduction}
With the development and integration of technologies such as diffusion~\citep{ho2020denoising, sohl2015deep}, multimodal~\citep{radford2021learning, ramesh2021zero}, pre-trained models~\citep{brown2020language}, and model fine-tuning~\citep{hu2021lora, li2021prefix}, artificial intelligence generated content (AIGC) has achieved remarkable progress, sparking widespread interest in interdisciplinary medical fields. 
Ranging from image reconstruction~\citep{liu2023dolce} and translation~\citep{wang2024soft} to the generation of virtual cases and the creation of simulation data~\citep{ozbey2023unsupervised}, AIGC underpin numerous innovations. 
This has significantly advanced the frontiers of computer-assisted diagnosis and precision medicine.
Recently, advances in video generation models~\citep{blattmann2023align, an2023latent, huang2024free} have propelled AIGC to new heights. 
However, due to the considerable complexity and significant resource demands of clinical video generation, research in this area remains in its nascent stages.
In this context, our work aims to explore whether it is possible to create temporally coherent and realistic clinical medical videos.

\begin{wrapfigure}{r}{8.5cm}
	\centering
	\includegraphics[width=8.5cm]{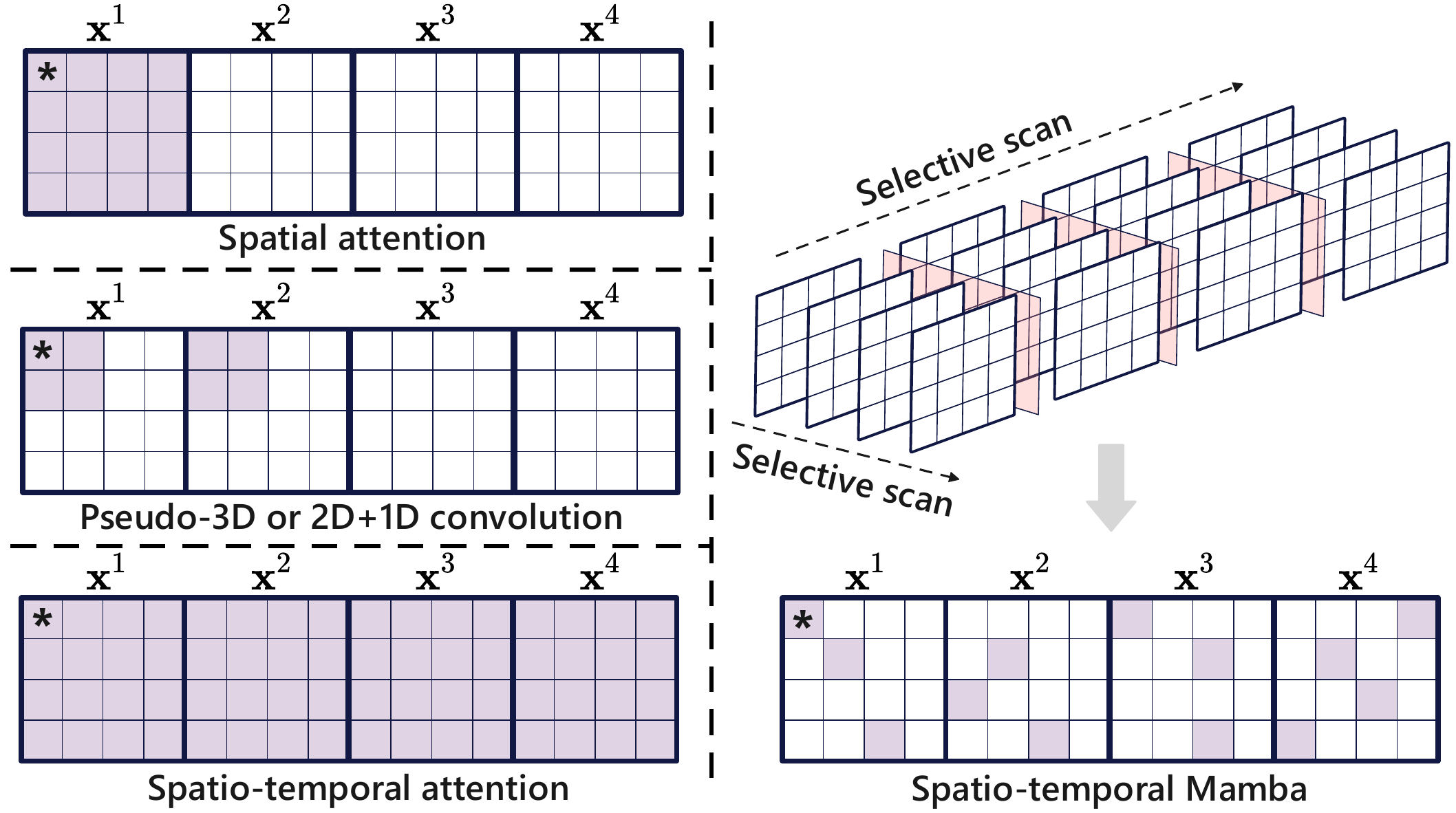}
	\vspace{-3mm}
	\caption{Illustration of spatial attention, pseudo-3D or 2D+1D convolution, spatio-temporal attention, and our spatio-temporal Mamba. 
		$\mathbf{x}^k$ represents the feature map of the $k$-th video frame.
		Patches marked with '*' are computed together with the colored patches to aggregate their features. }
	\label{fig:difference}
\end{wrapfigure}

Significant progress has been made in generating realistic medical images using generative adversarial networks (GANs)~\citep{goodfellow2014generative} and diffusion models~\citep{rombach2022high}. However, video is composed of pixel arrays spanning both temporal and spatial dimensions, necessitating meticulous attention to fine temporal dynamics and the maintenance of temporal consistency across frames. Consequently, generating stable, high-quality videos is a non-trivial task. 
To achieve this, most recent works in computer vision attempt to extend the spatial self-attention mechanism of images to spatio-temporal self-attention~\citep{khachatryan2023text2video, qi2023fatezero}, using pseudo-3D~\citep{singer2022make} or serial 2D+1D convolutions~\citep{xing2024make} to apply image diffusion models to video generation. 
To the best of our knowledge, the sole existing study in the medical field on video generation employs the spatio-temporal self-attention~\citep{li2024endora}.
Those strategies integrate features from patches across different video frames via an expanded attention or convolution module, as illustrated in Figure~\ref{fig:difference}. 
Despite their effectiveness in capturing contextual information across space and time, these methods have significant drawbacks: 
\textit{i)} spatio-temporal self-attention requires each patch attends to all other patches in the video, which not only poses the risk of misdirecting the attention mechanism with patches that are irrelevant to video generation~\citep{liu2022video}, but also significantly increases the computational complexity, presenting significant challenges in terms of resource allocation and maintaining cross-frame coherence,
and \textit{ii)} convolutional methods focus only on local areas and cannot capture the global receptive field, thereby reducing model performance.
To mitigate these challenges,
some recent works have incorporated Mamba~\citep{gu2023mamba, dao2024transformers} into video generation, they typically replace the original convolution or attention modules of GANs or diffusion models with Mamba~\citep{gao2024matten, mo2024scaling}, or merely alter the scanning order of patches~\citep{park2024videomamba} without thorough exploration.
These Mamba-based video generation models exhibit two primary limitations: 
\textit{i)} existence of pixel-level bias. Recent advances in diffusion models for image generation illustrate that Mamba underperforms relative to certain Transformer in generating pixel-level data~\citep{wang2024soft}. 
And \textit{ii)} absence of explicit spatio-temporal modeling. These models either maintain the same structural design of image generation models~\citep{mo2024scaling, gao2024matten} or simplistically modify the sequence of video frames~\citep{park2024videomamba}, assuming it to suffice for spatio-temporal modeling.
\textit{We aim to address these challenges by advancing our approach with an optimized design of spatio-temporal Mamba and attention blocks.}

Spatio-temporal self-attention frameworks typically model the content within each video frame before conducting temporal modeling of patches at corresponding positions across frames~\citep{diba2023spatio}. 
However, this approach has a significant limitation as it presupposes linear motion of patches along the temporal axis.
This assumption leads to inadequate mutual attention among patches following non-linear trajectories across frames.
\textit{To overcome existing drawbacks, we propose a supervised, optical flow representation alignment method seamlessly integrated into the video diffusion model.} By using optical flow~\citep{fleet2006optical, shi2023videoflow} to implicitly control key features, our method addresses the visual consistency problems identified in previous work. An primary benefit is the facilitation of precise information transfer between frames through optical flow guidance, stabilizing the generated visual content throughout the video.
Specifically, we initially employ a pre-trained optical flow estimation model~\citep{teed2020raft} to determine the optical flow of the source video. 
Following this, a self-supervised foundation model~\citep{caron2021emerging} is utilized to extract key features of the optical flow from various perspectives. These key features are then sampled and leveraged to supervise the training of video diffusion model, without the addition of any extra training parameters.

Video generation models process high-dimensional videos as cubic arrays, necessitating substantial memory and computational resources, particularly for high-resolution and extended-duration videos.
To reduce the computational load, numerous video diffusion models~\citep{blattmann2023align, an2023latent} are not trained on raw pixels, but first employ an autoencoder~\citep{kingma2013auto} to project video frames into low-dimensional latent space and then model this latent distribution.
However, the pre-train time-agnostic image autoencoder can distorts the temporal information in the latent space~\citep{zhou2022magicvideo}, \ie, the change $\triangle _z={z_i\to z_j }$ from the $i$-th frame to the $j$-th frame in the latent space suffers from frame consistency degradation or misalignment compared to the change $\triangle _x={x_i\to x_j }$ in the original pixel space. 
To address this challenge, \cite{lin2023catch} using additional modules, and \cite{xiang2023versvideo} retrained the decoder for image autoencoder. 
Researchers have recently begun training video autoencoders on large-scale video datasets~\citep{yang2024cogvideox}. However, the limited availability of medical videos means that these video autoencoders have not been adequately trained on medical datasets.
\textit{We incorporate frequency compensation modules into the video VAE to preserve spatio-temporal consistency, particularly during the projection and reconstruction processes of medical videos under spectral constraints.}

The primary contributions of this paper are: 
\textit{i)} proposing a novel video diffusion model framework that introduces spatio-temporal capabilities while minimizing resource requirements, featuring powerful spatio-temporal Mamba module and local attention mechanism.
\textit{ii)} optical flow representation alignment is incorporated into the video diffusion model to improve temporal coherence and smoothness of the generated videos.
\textit{iii)} development of a medical video autoencoder incorporating frequency compensation to enhance the reconstruction quality of medical videos.
Comprehensive experiments verify that our method \modell establishes a new benchmark in state-of-the-art performance among existing medical video generation models, particularly in sustaining visual consistency and diminishing computational burden.

\section{Preliminaries}
\subsection{Latent Diffusion Model}
Latent Diffusion Models (LDM)~\citep{rombach2022high} are probabilistic denoising models that integrate autoencoders with predefined posterior distributions, rapidly becoming foundational components in the AIGC field. These models consist of two main processes: diffusion and denoising. During the diffusion process, random noise is progressively added to the latent representation $\mathbf{z}$ through a sequence governed by a Markov chain with T-steps.
Specifically, consider a fixed $\mathbf{z}$ and a predetermined signal-noise schedule $\{\alpha_t, \sigma_t\}$, where the signal-to-noise ratio (SNR), given by $(\alpha_t^2/\sigma_t^2)$, decreases monotonically as $t$ increases. Define
$\alpha_{t|s}=\alpha_{t}/\alpha_{s}$ and $\sigma^2_{t|s}=\sigma^2_{t}-\alpha^2_{t|s}\sigma^2_{s}$ for all $s < t$,
we specify a series of latent variables $\mathbf{z}_t$ for $t = 0, \cdots, T$, which adhere to the following specified conditions:
\begin{eqnarray}\label{eq:eq1}
q(\mathbf{z}_t |\mathbf{z}_s )=
\begin{cases}
	\mathcal{N} (\mathbf{z}_t ; \alpha_t\mathbf{z}_s ,\sigma^2_t \mathbf{I} ) & \text{ if } s=0 \\
	\mathcal{N}(\mathbf{z}_t ; \alpha_{t|s}\mathbf{z}_s ,\sigma^2_{t|s} \mathbf{I} )  & \text{ if } s>0
\end{cases},
\end{eqnarray}	
where $\mathbf{I}$ represents the identity matrix.
During the denoising phase, LDM train a reverse model $p_{\theta}(\mathbf{z} _s|\mathbf{z} _t) $, which reformulates a denoising objective,
\begin{eqnarray}\label{eq:eq2}
	\begin{split}
&p_{\theta}(\mathbf{z} _s|\mathbf{z} _t)=\mathcal{N}(\mathbf{z} _s;\mu_\theta(\mathbf{z} _t,\mathbf{c}, t), \Sigma_{\theta}(\mathbf{z} _t,\mathbf{c}, t) ), \\
	\end{split}
\end{eqnarray}	
here $\mathbf{c}$ represents the conditional input. 
The mean function $\mu_\theta$ and the covariance matrix $\Sigma_{\theta}$ of the conditional distribution in the inverse process are calculated by training a denoising model $\epsilon_{\theta}$.

\subsection{Scalar State-Space Models}
The state space consists of a minimal set of variables that fully describe every possible state of the system.
Building upon this concept, State-Space Models (SSM)~\citep{gu2021efficiently, gu2021combining, smith2022simplified} define these variables to represent the system's states and can predict future states based on specific inputs.
To summarize, SSM defines a mapping from $\mathbf{x}\in\mathbb{R}^{(T,P)}$ to $\mathbf{y}\in\mathbb{R}^{(T,P)}$. The state transition is:
\begin{eqnarray}\label{eq:eq3}
	\begin{split}
		\mathbf{h}_t =\mathbf{A}_t\mathbf{h}_{t-1} + \mathbf{B}_{t}\mathbf{x}_t, \quad 
		\mathbf{y}_t =\mathbf{C}_t^{\mathsf{T}}\mathbf{h}_t,
	\end{split}
\end{eqnarray}
where $\mathbf{B}_t, \mathbf{C}_t \in \mathbb{R}^{(T,N)}$. $\mathbf{h}_t$ is referred to as the hidden state, represented by an $N$-dimensional vector, where $N$ is an independent hyperparameter, commonly referred to as the state dimension, state expansion factor, or state size.
SSM are classified as regular (unstructured) SSM, diagonal (structured) SSM~\citep{gu2023modeling, gupta2022diagonal}, or scalar SSM (also known as SSD)\citep{dao2024transformers}, depending on whether the dimensions of $\mathbf{A}_t$ are $\mathbb{R}^{(T,N,N)}$, $\mathbb{R}^{(T,N)}$, or $\mathbb{R}^{(T)}$, respectively. 
Selective SSM permit these parameters to change over time. Mamba\citep{gu2023mamba}, specifically its core "S6" layer, exemplifies a selective SSM with diagonal structure. For computational efficiency, Eq.\ref{eq:eq3} is implemented using matrix multiplication as follows:
\begin{eqnarray}\label{eq:eq4}
	\begin{split}
		\mathbf{y}&=\mathrm{SSM}(\mathbf{A},\mathbf{B},\mathbf{C})(\mathbf{x})=\mathbf{Mx}, \\
		~~ \mathrm{s.t.} ~~
		\mathbf{M}_{ij}&=
		\begin{cases}
			\mathbf{C}_i^{\mathsf{T}}\mathbf{A}_{i:j}^{\mathsf{X}}\mathbf{B}_j := \mathbf{C}_i^{\mathsf{T}}\mathbf{A}_i \cdots \mathbf{A}_{j+1}\mathbf{B}_j& \text{ if } i \ge j
			\\
			0 &\text{ if } i < j
		\end{cases}.
	\end{split}
\end{eqnarray}			
Here $\mathbf{M}$ is identified as a lower triangular semiseparable matrix. Mamba-2~\citep{dao2024transformers} imposes a constraint that all diagonal elements of $\mathbf{M}$ are identical, thereby restricting the diagonal to a scalar multiple of the identity matrix.

\subsection{Optical Flow Estimation}
The concept of optical flow~\citep{zhai2021optical}, originally introduced by Gibson~\citep{gibson1950perception}, represents a dense displacement vector field that describes the motion of pixels between consecutive frames under ideal conditions.
Based on the displacement vector field $(u, v)$, the coordinates $(\mathbf{x}^k_x, \mathbf{x}^k_y)$ of each pixel in the $k$-th frame can be projected to the corresponding coordinates in the ($k+1$)-th frame.
In the ($k+1$)-th frame, the new coordinates are determined by the coordinates in the $k$-th frame and the displacement vector field,
\begin{eqnarray}\label{eq:eq5}
	\begin{split}
		(\mathbf{x}^{k+1}_x, \mathbf{x}^{k+1}_y) = (\mathbf{x}^{k}_x + u^k(\mathbf{x}^{k}_x, \mathbf{x}^{k}_y), \mathbf{x}^{k}_y + v^k(\mathbf{x}^{k}_x, \mathbf{x}^{k}_y)).
	\end{split}
\end{eqnarray}	
Optical flow provides motion information that aids in understanding the dynamic and static regions within a moving scene. It accurately tracks the paths of objects in the video and effectively compensates for image instability caused by camera shaking or vibration, thereby ensuring smoother video presentation.

\section{Methodology}
Consider a dataset of videos $\mathcal{D}$, where each sample $\mathbf{x}^{1:F}$ is drawn from an unknown data distribution $p_{\mathrm{data}}(\mathbf{x}^{1:F})$. In this context, each $\mathbf{x}^{1:F} := (\mathbf{x}^1, \cdots, \mathbf{x}^F)$ corresponds to a sequence of video frames, with each frame sequence having a length $F>1$ and a resolution of $H\times W$. Specifically, each frame $\mathbf{x}^{\ell} \in \mathbb{R}^{(C,H,W)}$, where $C$ denotes the number of channels. The goal is to learn a model distribution $p_{\mathrm{model}}(\mathbf{x}^{1:F})$ that precisely matches the original data distribution $p_{\mathrm{data}}(\mathbf{x}^{1:F})$.

\begin{figure}[tbp]
	\label{fig:framework}
	\centering
	\includegraphics[width=0.99\linewidth]{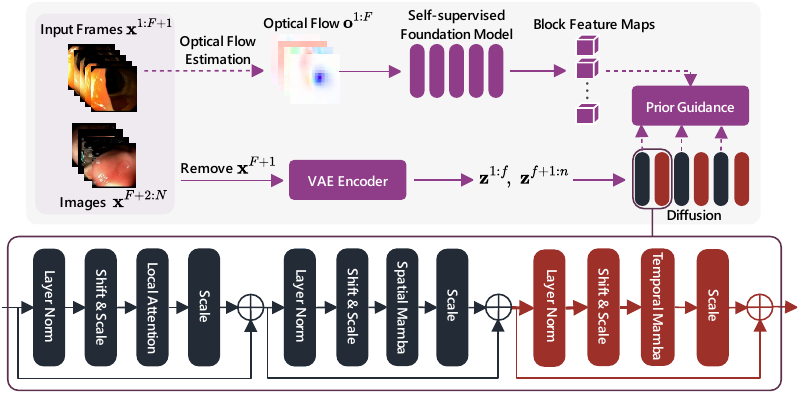}
	\caption{
		\modell is built upon the video Mamba diffusion model, the optical flow representation alignment component, and the frequency compensation video VAE. The Video Mamba diffusion model enables video generation under low resource load; the optical flow representation alignment aims to produce smoother videos and accelerate model convergence; the frequency compensation video VAE addresses potential inconsistencies during the video reconstruction process.}
	\label{fig:MedSora_Guidence_1}
\end{figure}

In the ensuing sections, we commence by detailing the architecture of diffusion model employed in \model, with a particular emphasis on  components spatio Mamba and temporal Mamba (Section~\ref{sec:3.1}). 
Subsequently, we elucidate the implementation of the optical flow representation alignment (Section~\ref{sec:3.2}), followed by delving into the comprehensive fine-tuning strategy for the 3D causal VAE (Section~\ref{sec:3.3}).
Finally, we provide a image and video joint training strategy (Section~\ref{sec:3.4}).

\subsection{Video Mamba Diffusion Model} \label{sec:3.1}
\begin{figure}[tbp]
	\centering
	\includegraphics[width=\linewidth]{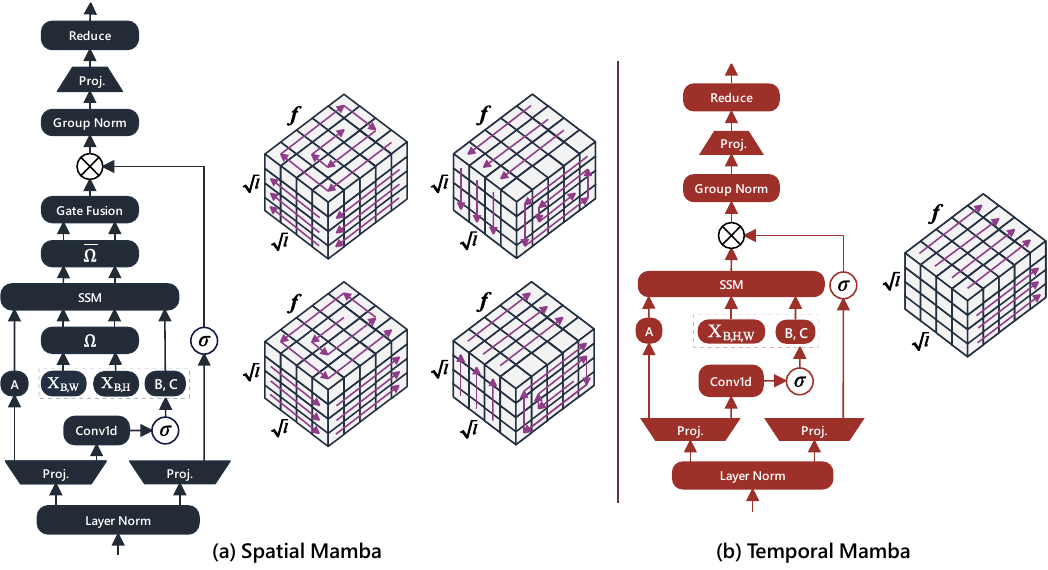}
	\caption{We present the structural details of spatial Mamba and temporal Mamba, both of which adopt the scalar state-space model. Spatial Mamba employ a bidirectional spiral scanning scheme to emphasize spatial continuity, while temporal Mamba scan along the frame axis.} 
	\label{fig:MedSora_Guidence_2}
\end{figure}

Current video diffusion models in the natural imaging typically employ either 3D or 2D+1D operations, simplifying the model excessively and thus compromising spatio-temporal performance. Alternatively, using standard self-attention mechanisms leads to significant computational overhead.
In Figure~\ref{fig:MedSora_Guidence_1}, each block of our video diffusion framework consists of modules sequentially connected via spatial and temporal components.
Considering Mamba's limitations in local modeling and its proficiency in handling long sequences, and given the self-attention mechanism's excellent local modeling capabilities but high computational cost that grows quadratically with sequence length, we propose employing local attention within the spatial component to model local information.
To expand the receptive field to cover an entire video frame, we use Mamba to model long sequences along both the width and height axes, as shown in Figure~\ref{fig:MedSora_Guidence_2}.
\textit{The motivation for this is that both the width and height axes are high-dimensional, and correlations generally weaken with increasing distance, prompting us to selectively scan along these axes.}

Given a batch of video latent $\mathbf{z}^{1:f}\in\mathbb{R}^{(b, f, l, d)}$, derived from a batch of original videos $\mathbf{x}^{1:F}\in\mathbb{R}^{(B, F, C ,H, W)}$ via a VAE, where $l$ denotes the number of tokens in a sequence and $d$ is the dimension of each token. The factorized local attention branch of the spatial component initially reshapes $\mathbf{z}^{1:f}$ into $(bf, l, d)$ to facilitate spatial attention.
For the $i$-th token in a sequence, define a sliding window $S(i)$ that encompasses the current position along with surrounding positions,
\begin{eqnarray}\label{eq:eq6}
	\begin{split}
		S(i) =\{j ~| ~j\in[\max(1,i-\left \lfloor \frac{{w}}{2}  \right \rfloor ), \min (f,i+\left \lfloor \frac{{w}}{2} \right \rfloor )]\}, 
	\end{split}
\end{eqnarray}	
here ${w}$ denotes the window size. The token at position $i$ is then processed through multi-head attention with other tokens within the window to determine the output for that position,
\begin{eqnarray}\label{eq:eq7}
	\begin{split}
		& \quad\quad \mathrm{\mathbf{Scores}} _{ij}^{{h} }  = \frac{\mathbf{Q}_i^{{h} }(\mathbf{K}_j^{{h} })^\mathsf{T}}{\sqrt{d_k} }, ~~  j\in S(i),~~ {h}\in \bar{H}, \quad\quad\quad
		 \alpha_{ij}^{{h}} = \frac{\exp (\mathrm{\mathbf{Scores}} _{ij}^{{h}})}{ {\textstyle \sum_{k\in S(i)}}\exp(\mathrm{\mathbf{Scores}} _{ik}^{{h}}) } , \\
		& \mathrm{\mathbf{Attention}}_i^{{h}} =\!\!\! \sum_{j \in S(i)} \!\!\alpha _{ij}^{{h}} \mathbf{V}_j^{{h}} ,  \quad
		\mathrm{\mathbf{Attention}}_i \!=\! \mathrm{Concat}(\mathrm{\mathbf{Attention}}_i^1, \cdots, \mathrm{\mathbf{Attention}}_i^{\bar{H}})\bar{W}_o,
	\end{split}
\end{eqnarray}	
where $\bar{H}$ is the number of heads in multi-head attention. $\mathbf{Q}^{{h}}=\mathbf{z}^{1:f}\bar{W}_q^{{h}}$, $\mathbf{K}^{{h}}=\mathbf{z}^{1:f}\bar{W}_k^{{h}}$, $\mathbf{V}^{{h}}=\mathbf{z}^{1:f}\bar{W}_v^{{h}}$,  and each of $\bar{W}_q^{{h}}$, $\bar{W}_k^{{h}}$, $\bar{W}_v^{{h}}$, $\bar{W}_o$ is a trainable weight matrix.
We expand the input tensor $\mathbf{z}^{1:f}$ along the width and height axes to perform spatial Mamba in the shapes $(b\bar{h}, f\bar{w}, d)$ and $(b\bar{w}, f\bar{h}, d)$, respectively, where $\bar{w}=\bar{h}=\sqrt{l}$. To fuse the selective scanning results from these two directions, we integrate them using a gating mechanism.

As shown in Figure~\ref{fig:MedSora_Guidence_2}, prior to performing the selective scan with Mamba, the sequence of $f\bar{w}$ or $f\bar{h}$ tokens is rearranged according to the spiral direction and its reverse (denoted as $\Omega$). 
Selective scan are then performed on both rearranged sequences. Subsequently, the original order of the results is restored using the spiral indexes, and the corresponding tokens at each position are summed (denoted as $\bar{\Omega}$). 
Without loss of generality, given an input tensor of shape $(\hat{b}, \hat{l}, d)$, where $\hat{b}=b\bar{h}$, $\hat{l}=f\bar{w}$ for height axis scanning or $\hat{b}=b\bar{w}$, $\hat{l}=f\bar{h}$ for width axis scanning. 
The spiral scanning process is
\begin{eqnarray}\label{eq:eq8}
	\begin{split}
		&\{\mathbf{z}^{1:f}_{1},\mathbf{z}^{1:f}_{2}, \cdots,\mathbf{z}^{1:f}_{\hat{l}}\}\xrightarrow{\mathrm{rearrange~per~}\Omega} \{\mathbf{z}^{1:f}_{\Omega_1},\mathbf{z}^{1:f}_{\Omega_2}, \cdots,\mathbf{z}^{1:f}_{\Omega_{\hat{l}}}\}, \{\mathbf{z}^{1:f}_{\Omega_{\hat{l}}}, \mathbf{z}^{1:f}_{\Omega_{\hat{l}-1}}, \cdots, \mathbf{z}^{1:f}_{\Omega_1}\} , \\
		&\{\bar{\mathbf{y}}_{\Omega_1}, \bar{\mathbf{y}}_{\Omega_2}, \cdots, \bar{\mathbf{y}}_{\Omega_{\hat{l}}}\} = \mathrm{SSM}({\mathbf{A}}, {\mathbf{B}}, \mathbf{C})(\{\mathbf{z}^{1:f}_{\Omega_1},\mathbf{z}^{1:f}_{\Omega_2}, \cdots,\mathbf{z}^{1:f}_{\Omega_{\hat{l}}}\} ) ,\\
		&\{\hat{\mathbf{y}}_{\Omega_{\hat{l}}}, \hat{\mathbf{y}}_{\Omega_{\hat{l}-1}}, \cdots, \hat{\mathbf{y}}_{\Omega_1}\} = \mathrm{SSM}({\mathbf{A}}, {\mathbf{B}}, \mathbf{C})(\{\mathbf{z}^{1:f}_{\Omega_{\hat{l}}},\mathbf{z}^{1:f}_{\Omega_{\hat{l}-1}}, \cdots,\mathbf{z}^{1:f}_{\Omega_1}\} ) ,\\
		&\{\bar{\mathbf{y}}_{\Omega_1}, \bar{\mathbf{y}}_{\Omega_2}, \cdots, \bar{\mathbf{y}}_{\Omega_{\hat{l}}}\}, \{\hat{\mathbf{y}}_{\Omega_{\hat{l}}}, \hat{\mathbf{y}}_{\Omega_{\hat{l}-1}}, \cdots, \hat{\mathbf{y}}_{\Omega_1}\} \xrightarrow{\mathrm{rearrange~per~}\bar{\Omega}}  \{{\mathbf{y}}_{1}, {\mathbf{y}}_{2}, \cdots, {\mathbf{y}}_{{\hat{l}}}\} .
	\end{split}
\end{eqnarray}
Here $\mathbf{z}^{1:f}_{i}\in \mathbb{R}^{(\hat{b}, 1, d)}$ denotes the $i$-th token in the sequence. Following a spiral scan across both width and height dimensions, gated fusion is utilized to synthesize the outcomes from both axes,
\begin{eqnarray}\label{eq:eq9}
	\begin{split}
		\mathbf{y}=\sigma (W_1\mathbf{y}_w+b_1)\odot \mathbf{y}_w + \sigma (W_2\mathbf{y}_h+b_2)\odot \mathbf{y}_h,
	\end{split}
\end{eqnarray}
where $\mathbf{y}_w := \{{\mathbf{y}}_{1}, {\mathbf{y}}_{2}, \cdots, {\mathbf{y}}_{{f\bar{w}}}\}$ and $\mathbf{y}_h:=\{{\mathbf{y}}_{1}, {\mathbf{y}}_{2}, \cdots, {\mathbf{y}}_{{f\bar{h}}}\}$. 
$\sigma$ is activation function.
$W_1$, $W_2$, $b_1$ and $b_2$ are trainable parameters.

Following the modeling of individual video frames, we reshape the tensor $\mathbf{z}^{1:f}$ to shape $(bl,f,d)$ and apply the temporal Mamba to selective scan along the frame axis, thereby extending the receptive field to encompass the entire video.
\begin{eqnarray}\label{eq:eq10}
	\begin{split}	
		\{{\mathbf{y}}_1, {\mathbf{y}}_2,\cdots,{\mathbf{y}}_f\} = \mathrm{SSM}(\mathbf{A},\mathbf{B},\mathbf{C})(\{ \mathbf{z}^{1:f}_1, \mathbf{z}^{1:f}_2, \cdots,\mathbf{z}^{1:f}_f   \}).
	\end{split}
\end{eqnarray}

\textbf{Computation Efficiency.} For a video latent representation $\mathbf{z}^{1:f}\in\mathbb{R}^{(1,f,l,d)}$, the computational complexities of spatio-temporal self-attention and our method are as follows:
\begin{eqnarray}\label{eq:eq11}
	\begin{split}	
		&O(\text{Spatio-temporal~self-attention}) = \!\underbrace{f(4ld^2\!+\!2l^2d)}_{\text{spatial~self-attention}}\!+\! \underbrace{l(4fd^2\!+\!2f^2d)}_{\text{temporal~self-attention}}\!=\!8fld^2\!+\!2fl^2d\!+\!2lf^2d, \\
		&O(\text{Ours}) = \underbrace{f(4ld^2\!+\!2lwd)}_{\text{local~attention}}+\underbrace{2\bar{h}(3f\bar{w}(2d)\bar{N} \!+\! f\bar{w}(2d)\bar{N}^2)+2\bar{w}(3f\bar{h}(2d)\bar{N}\!+\!f\bar{h}(2d)\bar{N}^2)}_{\text{spatial Mamba}}\\
		&\quad\quad\quad~~+\underbrace{2l(2f(2d)\bar{N}+f(2d)\bar{N}^2)}_{\text{temporal Mamba}}
		=4fld^2+2flwd+18fld\bar{N}+6fld\bar{N}^2.
	\end{split}
\end{eqnarray}
$\bar{N}$ is a fixed parameter of Mamba, which is generally less than $l$. Eq.(\ref{eq:eq11}) demonstrates that the proposed video Mamba diffusion model significantly reduces the computational complexity.

\subsection{Optical Flow Representation Alignment} \label{sec:3.2}
Recent studies~\citep{chen2401deconstructing, xiang2023denoising, li2023your} have demonstrated that meaningful discriminative feature representations can be induced within diffusion models during the denoising process, achieved by implicitly learning the representation $\mathbf{h}$ as the hidden state of a denoising autoencoder that reconstructs the original data $\mathbf{x}$ from its corrupted version. 
Compared to relying solely on the diffusion model to independently learn these representations, this method can enhance training efficiency and improve the quality of generation~\citep{yu2024representation}.
This insight motivate us to enhance generative models by integrating external self-supervised representations.
However, applying this approach to video generation presents challenges. First, the complex pixel movements between video frames are not fully considered. Most existing studies focus on image generation tasks, and the few that address video generation~\citep{li2024endora} usually treat videos as sequences of individual images—inputting single-frame images into self-supervised models to obtain representations, and then integrating these representations.
Second, there is an input mismatch problem: most self-supervised learning encoders are trained on clean images, whereas latent diffusion models typically use compressed latent representations from VAE with added noise as input.
\textit{To address these problems, we start by using a pre-trained optical flow prediction model RAFT~\citep{teed2020raft} to track pixel motion across video frames, and employ a self-supervised visual model DINO~\citep{caron2021emerging} to extract motion feature representations. We then sample these motion feature representations and compute the covariance with the output of the spatial modeling from the video Mamba diffusion model, without requiring any additional parameter training.}

Optical flow estimation relies on analyzing consecutive frame pairs. To maintain consistency, we supply the optical flow prediction model with an extra frame, resulting in one more input frame than the VAE receives.
Initially, the input $\mathbf{x}^{1:F+1}$ is processed using the optical flow prediction model RAFT to generate optical flow vector field $\mathbf{O}^{1:F}$. The $i$-th field encompasses the horizontal component $\mathbf{O}^{i}_u$ and the vertical component $\mathbf{O}^{i}_v$, which define the vector transitions from the $i$-th frame to the $(i+1)$-th frame and can be expressed as
\begin{eqnarray}\label{eq:eq12}
	\begin{split}
		\mathbf{O}_u^{i}\!=\!
		{\small \begin{bmatrix}
			u^i(1,1)&u^i(1,2)  &\cdots  &u^i(1,W) \\
			u^i(2,1)&u^i(2,2)  &\cdots  &u^i(2,W) \\
			\vdots & \vdots & \ddots  &\vdots \\
			u^i(H,1)&u^i(H,2)  &\cdots  &u^i(H,W)
		\end{bmatrix}},
		\mathbf{O}_v^{i} \!=\!
		{\small \begin{bmatrix}
			v^i(1,1)&v^i(1,2)  &\cdots  &v^i(1,W) \\
			v^i(2,1)&v^i(2,2)  &\cdots  &v^i(2,W) \\
			\vdots & \vdots & \ddots  &\vdots \\
			v^i(H,1)&v^i(H,2)  &\cdots  &v^i(H,W)
		\end{bmatrix}}. 
	\end{split}
\end{eqnarray}
Since the self-supervised foundation model DINO accepts a three-dimensional RGB image as input, the two-dimensional optical flow field must be further processed.
Specifically, for the horizontal component $u^i(x,y)$ and vertical component $v^i(x,y)$ at position $(x, y)$ in the $i$-th optical flow field, we encode the two-dimensional optical flow vector into an RGB pixel. This representation not only intuitively illustrates the direction of motion but also effectively conveys the intensity of motion. Mathematically,
\begin{eqnarray}\label{eq:eq13}
	\begin{split}
	&M^i(x,y)=\sqrt{u^i(x,y)^2+v^i(x,y)^2}, \quad\quad~~~
	A^i(x,y)=\arctan \left(\frac{v^i(x,y)}{u^i(x,y)} \right) , \\
	&H^i(x,y)=\frac{A^i(x,y)}{2\pi } \times 360^{\circ }, \quad\quad S^{i}(x,y)=1, \quad\quad V^i(x,y)=\frac{M^i(x,y)}{M^i_{\max }}, \\
	&(R^i(x,y), G^i(x,y), B^i(x,y)) = \mathrm{HSVtoRGB}(	H^i(x,y), S^{i}(x,y), V^i(x,y)).
	\end{split}
\end{eqnarray}
Eq.(\ref{eq:eq13}) initially represents $u^i(x,y)$ and $v^i(x,y)$ in terms of motion intensity  $M^i(x,y)$ and direction  $A^i(x,y)$. 
Subsequently, these components are mapped to the HSV color space, where $H^i(x,y)$, $S^i(x,y)$, $V^i(x,y)$ respectively denote hue, saturation, and value.
Finally, the HSV pixel is converted to an RGB pixel using a standard color space conversion method.

We employ the self-supervised foundation model DINO, which exhibits high semantic relevance, to extract dense features from optical flow images. Specifically, the optical flow image derived from Eq.(\ref{eq:eq13}) is fed into the DINO model to yield multi-scale representations that span from shallow to deep layers, denoted as $[\mathbf{h}_1^*,\mathbf{h}_2^*,\cdots,\mathbf{h}_D^*]$. These representations are then aligned with the reshaped output features $[\mathbf{y}_1^*,\mathbf{y}_2^*,\cdots,\mathbf{y}_L^*]$ of the spatial component. Here $\mathbf{h}_i^* \in \mathbb{R}^{(\tilde{N},\tilde{K})}$, $\mathbf{y}_i^* \in \mathbb{R}^{(K,d)}$, $\tilde{K}=bF\tilde{l}$, $K=bfl$ and $\tilde{l}$ is the sequence length of DINO.
We utilize the relative distribution similarity~\citep{tang2024all} to match features,
\begin{eqnarray}\label{eq:eq14}
	\begin{split}
&\mathbf{r}_{i,d_i}^{1:f} \!:=
1\!-\!\frac{ {\textstyle \sum_{k=1}^{K}}(\mathbf{y}^*_{i,k,d_i} \!-\! \bar{\mathbf{y}}^*_{i,d_i}) (\mathrm{Sampling}(\mathbf{h}^*_{\nu_i,k,d_i}) \!-\! \bar{\mathbf{h}}^*_{\nu_i,d_i}) }
{\sqrt{{\textstyle \sum_{k=1}^{K}} (\mathbf{y}^*_{i,k,d_i} \!-\! \bar{\mathbf{y}}^*_{i,d_i})^2} \sqrt{{\textstyle \sum_{k=1}^{K}} (\mathrm{Sampling}(\mathbf{h}^*_{\nu_i,k,d_i}) \!-\! \bar{\mathbf{h}}^*_{\nu_i,d_i})^2} },~~ i \in L, d_i \in d, \\
&\quad\quad\quad\mathrm{where} \quad\quad \bar{\mathbf{h}}^*_{\nu_i,d_i} = \frac{1}{K}\sum_{k=1}^{K}\mathrm{Sampling}(\mathbf{h}^*_{\nu_i,k,d_i}), \quad \quad \bar{\mathbf{y}}^*_{i,d_i} = \frac{1}{K}\sum_{k=1}^{K}\mathbf{y}^*_{i,k,d_i}.
	\end{split}
\end{eqnarray}
We use $\mathbf{h}^*_{\nu_i}$ represents the optical flow representation corresponding to $\mathbf{y}^*_i$, which is selected from the set $[\mathbf{h}_1^*,\mathbf{h}_2^*,\cdots,\mathbf{h}_D^*]$. $\mathrm{Sampling(\cdot)}$ denotes the application of bilinear interpolation to the first dimension of $\mathbf{h}^*_{\nu_i}$ and $1 \times 1$ convolution to its second dimension, ensuring that $\mathbf{h}^*_{\nu_i}$ and 
$\mathbf{y}^*_{i}$ have consistent dimensions.
We define the optical flow representation alignment loss as follows:
\begin{eqnarray}\label{eq:eq15}
	\begin{split}
		\mathcal{L}_{\text{Align}}^{1:f} = \max (\frac{1}{Ld}\sum_{i=1}^{L} \sum_{d_i=1}^{d} g({t})\cdot \mathbf{r}_{i,d_i}^{1:f} - \mathrm{margin}, 0).
	\end{split}
\end{eqnarray}
$t\in[0, T)$ signifies the time step or stage index, and $T$ represents the predetermined total number of steps.
As ${t}$ progresses, the noise superimposed on the video latent intensifies, 
consequently diminishing the discriminative representational capacity of the output features from the diffusion model's spatial components. 
In response, we propose introducing a hyperparameter $g({t})$, which is inversely proportional to ${t}$,
to regulate the variable $\mathbf{r}_{i,d_i}^{1:f}$, set to ${1}/{(t+1)}$ by default.
Furthermore, we incorporate a hyperparameter $\mathrm{margin}$ in Eq.(\ref{eq:eq15}) for regularization purposes, aiming to enhance the separation or decorrelation among features.
By neglecting small $\mathbf{r}_{i,d_i}^{1:f}$, we prevent the model from overemphasizing minor differences, thereby steering the training process towards correcting more substantial errors and enhancing the model's robustness.

\subsection{Frequency Compensation Video VAE} \label{sec:3.3}
Many video diffusion models employ image VAEs to compress source videos into latent space and then reconstruct the videos within that space~\citep{wu2023tune, chen2023control, xing2024simda}.
However, since videos encapsulate both spatial and temporal information, the absence of temporal components in image VAEs precludes capturing inter-frame dependencies, often resulting in artifacts, flickering, or temporal inconsistencies~\citep{blattmann2023align, xiang2023versvideo}.
Furthermore, given that video data significantly exceeds image data in size, and image VAEs solely compress spatial dimensions, they fail to efficiently compress along the temporal dimension, resulting in relatively large latent representations for videos.

\begin{figure}[tbp]
	\centering
	\includegraphics[width=\linewidth]{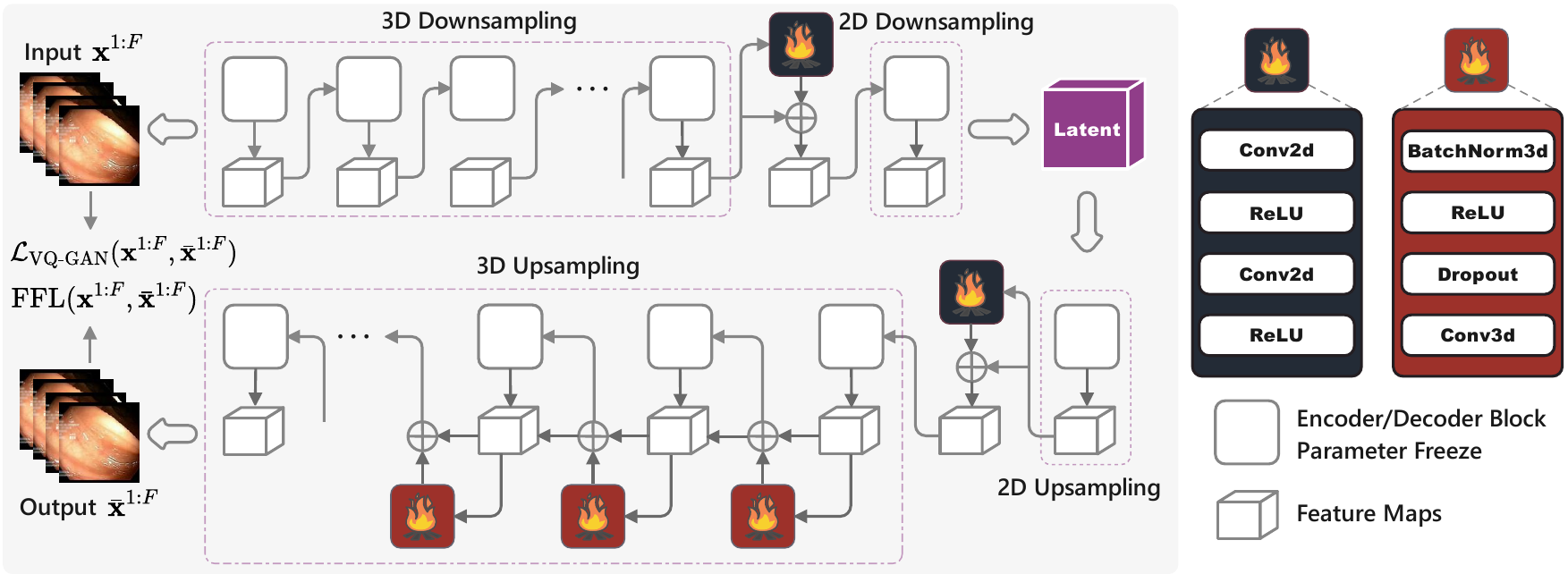}
	\caption{The proposed frequency compensation video VAE. Current video diffusion models either rely on image VAEs for encoding and reconstructing videos, potentially omitting crucial temporal information and resulting in inadequate compression rates, or  rely on video VAEs that lack training on medical videos. The proposed frequency compensation video VAE introduces frequency compensation components with two distinct structures, built on the 3D causal VAE, ensuring temporal consistency while focusing on the structural information of medical videos, which is instrumental in identifying areas such as lesions and textures.}
	\label{fig:MedSora_VAE}
\end{figure}

Recent works have implemented 3D VAEs video compression module by employing 3D convolutions to compress both the temporal and spatial dimensions of videos~\citep{yang2024cogvideox}. 
However, the scarcity of medical video datasets often hinders these VAEs from being adequately trained, leading to insufficient generalization in medical field.

\textit{To effectively adapt to the medical video generation task, and in light of the significant advantages of frequency methods in processing medical data with complex structural features, we introduced frequency compensation modules into the 3D causal VAE~\citep{yang2024cogvideox}.}
During the training phase, in addition to employing the same optimization loss as VQ-GAN~\citep{esser2021taming},
we also introduced the Focal Frequency Loss~\citep{jiang2021focal} to narrow the gap between the original and generated videos in the frequency domain, thereby more accurately capturing structural information such as the edge and texture features of the lesion.
To optimize training efficiency, we freeze the original parameters of the 3D causal VAE and insert the frequency compensation modules
before the encoder's 2D downsampling block and in the middle of each decoder block.
As shown in Figure~\ref{fig:MedSora_VAE}.

The frequency compensation modules contain two components with different structures. The component close to the 2D sampling blocks uses standard 2D convolution to reconstruct space, and the other component uses standard 3D convolution to reconstruct time and space. 
The final optimization loss for fine-tuning the 3D causal VAE is formulated as follows:
\begin{eqnarray}\label{eq:eq16}
	\begin{split}
		\mathcal{L}_{\text{VAE}} = \mathcal{L}_{\text{VQ-GAN}}(\mathbf{x}^{1:F}, \bar{\mathbf{x}}^{1:F}) + \lambda \mathrm{FFL}(\mathbf{x}^{1:F}, \bar{\mathbf{x}}^{1:F}).
	\end{split}
\end{eqnarray}
$\mathcal{L}_{\text{VQ-GAN}}(\cdot, \cdot)$ is the reconstruction loss as in VQ-GAN, includes L1 loss and perceptual similarity loss~\citep{zhang2018unreasonable}, $\lambda$ is a weight coefficient, set to $0.1$ by default.

\subsection{Image and Video Joint Training} \label{sec:3.4}
Several studies have demonstrated that the joint training of videos and images significantly improves the quality and diversity of video generation~\citep{ho2022video, gupta2023photorealistic, wang2023modelscope} while mitigating the phenomenon of forgetting~\citep{chen2023videocrafter1}.
To implement this joint training, we randomly select independent images $\mathbf{x}^{F+2:N}$ from various videos within the same dataset and concatenate them to the end of each video $\mathbf{x}^{1:F}$ sampled randomly from the dataset.
The training loss for the video Mamba diffusion model is
\begin{eqnarray}\label{eq:eq17}
	\begin{split}
&\quad\quad\quad\quad\quad\quad   \mathcal{L}_{\text{DDPM}}  = \|\epsilon-\epsilon_\theta(\mathbf{z}_t^{1:f}, \mathbf{z}_t^{f+1:n},t)\|^2 + \alpha \mathcal{L}_{\text{Align}}^{1:f}, \\
& \mathrm{s.t.}\quad t\in[0,T), ~ \mathbf{z}_t^{1:f}\sim q(\mathbf{z}_t^{1:f}|\mathbf{z}_s^{1:f}), ~ \mathbf{z}_t^{f+1:n}\sim q(\mathbf{z}_t^{f+1:n}|\mathbf{z}_s^{f+1:n}), ~ \epsilon\sim \mathcal{N}(0,\mathbf{I}), \\
	\end{split}
\end{eqnarray}
where $\alpha$ is a hyperparameter. We mask the images when computing $\mathcal{L}_{\text{Align}}^{1:f}$ to prevent the introduction of spurious optical flow motion trajectories between individual images.

\section{Experiments}
\subsection{Implementation Details}
Video Mamba diffusion model was trained using the AdamW optimizer~\citep{loshchilov2017decoupled}, which employs a phased learning rate adjustment strategy. During the initial $2$ epochs, a learning rate of $1\times 10^{-3}$ 
and a batch size of $16$
was used to accelerate convergence. Subsequently, the learning rate was fixed at $1\times 10^{-4}$ and the batch size was set to $2$ to achieve more stable and fine-grained parameter optimization.
In alignment with the settings of the sole existing medical video generation work~\citep{li2024endora}, we configured the training video to consist of $16$ frames, each with a resolution of $128\times 128$ pixels. In addition, we randomly selected $8$ images for joint training, each matching the resolution of the video frame.
Data augmentation was performed solely through horizontal flipping. Consistent with standard practices in diffusive generative models, we did not modify the decay/warm-up schedule, learning rate, or Adam $\beta_1/\beta_2$ hyperparameters.
Training incorporated the exponential moving average (EMA)~\citep{tarvainen2017mean, wang2022metateacher} with a decay rate of $0.9999$. The performance outcomes of the EMA model were reported for final result sampling.
Each dataset undergoes training for one week using early stopping strategy to prevent overfitting, in line with standard practices for video-related tasks.
Gradient clipping is initiated at the $25$-th epoch, with the clip max norm set to $1.0$ to ensure training stability.
Unless otherwise specified, the hyperparameter $\alpha$ is set to $0.01$.
Most of our experiments utilized multiple Nvidia RTX $3090$ Ti GPUs. 
During the inference stage, videos are generated via DDIM sampling~\citep{song2020denoising}.
Additional training details and network specifications are outlined in Table~\ref{tab:1}.

During the training of the frequency compensation video VAE, we utilize the Adam optimizer, conducting training across each dataset for $10$ epochs, setting the learning rate to $2\times 10^{-7}$, and configuring the momentum parameters $\beta_1 $and $\beta_2$
to $0.5$ and $0.9$, respectively.

\begin{table}[t] 
	\caption{\modell network detail. We introduce three variants of our model, distinguished by network depth, selected DINO layer features, and parameter count. The depth and patch size can be adjusted to scale the network as desired. Notably, our model has significantly fewer parameters compared to existing video generation models of same depth and patch size.} \vspace{0.5em}
	\label{tab:1}
	\begin{tabular}{@{}llll@{}}
		\toprule
		& \textbf{MedSora-S}                   & \textbf{MedSora-B}                      & \textbf{MedSora-L}  \\                                                             \midrule
		\textbf{Architecture}           & \textbf{LDM}          & \textbf{LDM}          & \textbf{LDM}          \\
		video resolution                & $16\times 128\times 128\times 3$ & $16\times 128\times 128\times 3$ & $16\times 128\times 128\times 3$ \\
		images resolution               & $8\times 128\times 128\times 3$  & $8\times 128\times 128\times 3$  & $8\times 128\times 128\times 3$  \\
		z-shape                         & $6\times 16\times 16\times 16$   & $6\times 16\times 16\times 16$   & $6\times 16\times 16\times 16$   \\
		Embedding dimension             & $512$                            & $512$                            & $512$                            \\
		Depth                           & $6$                              & $12$                             & $24$                             \\
		Patch size                      & $2$                              & $2$                              & $2$                              \\
		Strip size                      & $2$                              & $2$                              & $2$                              \\       \midrule
		\textbf{Local attention}        &                                  &                                  &                                  \\
		Window size                     & $8$                              & $8$                              & $8$                              \\
		Look backward                   & $1$                              & $1$                              & $1$                              \\
		Look forward                    & $1$                              & $1$                              & $1$                              \\
		Dropout                         & $0.1$                            & $0.1$                            & $0.1$                            \\
		Exact window size               & $\mathrm{True}$                  & $\mathrm{True}$                  & $\mathrm{True}$                  \\       \midrule
		\textbf{Spatial$\&$Temporal Mamba} &                               &                                  &                                  \\
		State expansion factor          & $128$                            & $128$                            & $128$                            \\
		Local convolution width         & $4$                              & $4$                              & $4$                              \\
		Expand                          & $2$                              & $2$                              & $2$                              \\
		Head dimension                  & $64$                             & $64$                             & $64$                             \\
		A init range                    & $(1, 16)$                        & $(1, 16)$                        & $(1, 16)$                        \\
		Groups norm                     & $1$                              & $1$                              & $1$                              \\
	    $\Delta$ min                    & $0.001$                          & $0.001$                          & $0.001$                          \\
		$\Delta$ max                    & $0.1$                            & $0.1$                            & $0.1$                            \\
		$\Delta$ init floor             & $1\times 10^{-4}$                & $1\times 10^{-4}$                & $1\times 10^{-4}$                \\
		Chunk size                      & $256$                            & $256$                            & $256$                            \\       \midrule
		\textbf{Optical flow sampling}  &                                  &                                  &                                  \\
		Selected DINO Layers            &$[0,2,4,6,8,10]$                  & $0\to 11$                        & $0,0 \to 11,11$                   \\
		Bilinear interpolation size     & $(64, 512)$                      & $(64, 512)$                      & $(64, 512)$                      \\
		Convolution                     & $(1 \times 1, 16 \rightarrow 4)$ & $(1 \times 1, 16 \rightarrow 4)$ & $(1 \times 1, 16 \rightarrow 4)$ \\       \midrule
		\textbf{Training}               &                                  &                                  &                                  \\
		Parameterization                & $\epsilon$                       & $\epsilon$                       & $\epsilon$                       \\
		Learning rate                   & - & $1\!\times\!10^{-3}, ~1\!\times \! 10^{-4}$                &$1\!\times\!10^{-3}, ~1\!\times \! 10^{-4}$               \\
		Batch size                      & -                         & $16, 2$                          & $16, 2$                       \\
		Start gradient clipping         & -              & $25~\mathrm{epoch}$              & $25~\mathrm{epoch}$              \\
		Clip max norm                   & -                             & $1$                              & $1$                              \\
		Diffusion steps                 & -                           & $1000$                           & $1000$                           \\
		Noise schedule                  & $\mathrm{Linear}$                & $\mathrm{Linear}$                & $\mathrm{Linear}$                \\       \midrule
		\textbf{Sampling}               &                                  &                                  &                                  \\
		Sampler                         & $\mathrm{DDIM}$                  & $\mathrm{DDIM}$                  & $\mathrm{DDIM}$                  \\
		Steps                           & $250$                            & $250$                            & $250$                            \\       \midrule
		\textbf{Total training parameters}     &$43.76~\mathrm{M}$         & $86.47~\mathrm{M}$               & $171.89~\mathrm{M}$              \\  
		\bottomrule
	\end{tabular}
\end{table}

\subsection{Dataset Setting}
\textbf{Colonoscopic}~\citep{mesejo2016computer}. The Colonoscopic dataset comprises 210 colonoscopy videos, each annotated by medical experts to identify various types of lesions including polyps, cancers, and other non-malignant formations. Each video ranges from $30$ seconds to one minute in duration and includes $30$ frames per second, with each frame having a resolution of $768\times 576$ pixels. 

\textbf{CholecTriplet}~\citep{nwoye2022rendezvous}. The CholecTriplet dataset is an endoscopic video dataset that extends the CholecT40 dataset~\citep{nwoye2020recognition} for laparoscopic cholecystectomy surgery, annotated with triple information in the format ⟨instrument, verb, target⟩. The dataset comprises $50$ videos, including $45$ from the Cholec80 dataset~\citep{twinanda2016endonet} and $5$ from an internal dataset of the same surgical procedure.

\textbf{Kvasir-Capsul}~\citep{borgli2020hyperkvasir}. The Kvasir-Capsule dataset comprises $117$ videos, each annotated at the frame level. These videos were captured using capsule endoscope, a small swallowable camera device that can capture detailed videos of the gastrointestinal tract, particularly the small intestine. The dataset includes annotated videos across various gastrointestinal conditions, including normal findings and common illnesses, providing a valuable resource for the research and development of automated lesion detection algorithms.

We extracted 16 frame video clips from these datasets \textsl{by sampling every third frame,} and resized each frame to $128\times 128$ for training.

\subsection{Evaluation}
We assessed \modell employing five metrics: Fréchet Video Distance (FVD)~\citep{unterthiner2018towards}; Content-Debiased Fréchet Video Distance (CD-FVD)~\citep{ge2024content}, which is an enhanced version of the FVD, offering superior accuracy in measuring actual motion; Fréchet Inception Distance (FID)~\citep{heusel2017gans}; 
Inception Score (IS)~\citep{saito2017temporal}. And in our analysis study, we additionally used Frame Consistency (FC) to evaluate video continuity by computing the mean CLIP similarity between consecutive frames~\citep{esser2023structure}.
Following the evaluation criteria outlined in StyleGAN-V~\citep{skorokhodov2022stylegan}, we used $2048$ video clips to calculate the metrics.

\subsection{Quantitative Evaluation}

We compare our method \modell with four publicly available video generation methods: StyleGAN-V~\citep{skorokhodov2022stylegan}, MoStGAN-V~\citep{shen2023mostgan}, LVDM~\citep{he2022latent}, and Endora~\citep{li2024endora}. As presented in Table~\ref{tab:2}, \modell outperforms the state-of-the-art GAN-based methods StyleGAN-V and MoStGAN-V across all performance metrics. Furthermore, \modell surpasses the diffusion-based methods LVDM and Endora (base size) while utilizing fewer training parameters.

\begin{table}[htbp] \small
	\centering
	\caption{Quantitative comparisons between \modell and other video generation methods.} \vspace{0.5em}
	\label{tab:2}
	\begin{tabularx}{\textwidth}{@{}l C c *{2}{C}  C c *{2}{C}  C c *{2}{C}}
		\toprule
		&  \multicolumn{4}{c}{Colonoscopic} & \multicolumn{4}{c}{CholecTriplet} & \multicolumn{4}{c}{Kvasir-Capsul} \\ 
		\cmidrule(lr){2-5}  \cmidrule(lr){6-9} \cmidrule(lr){10-13}
		&  $\!\!\!$FVD$\downarrow$ & $~$CD-FVD$\downarrow$ & $\!\!\!\!\!$FID$\downarrow$ & $\!\!$IS$\uparrow$  & FVD$\downarrow$ & $~~~$CD-FVD$\downarrow$ & $\!\!\!\!\!\!$FID$\downarrow$ & $\!\!$IS$\uparrow$  & FVD$\downarrow$ & $~~~$CD-FVD$\downarrow$ & $\!\!\!\!\!\!$FID$\downarrow$ & $\!\!$IS$\uparrow$  \\
		StyleGAN-V  &$\!\!\!\!\!\!$\colorbox{top6}{\makebox[0.9cm][c]{1994.1}} &$\!\!\!\!$\colorbox{top6}{\makebox[0.9cm][c]{630.87}}  &$\!\!\!\!\!\!\!\!\!\!\!\!\!$\colorbox{top6}{\makebox[0.9cm][c]{226.59}} &$\!\!\!\!\!\!$\colorbox{top5}{\makebox[0.6cm][c]{1.95}} &$\!\!\!$\colorbox{top5}{\makebox[0.9cm][c]{648.37}} &$\!$\colorbox{top5}{\makebox[0.9cm][c]{747.35}} &$\!\!\!\!\!\!\!\!\!\!\!\!\!$\colorbox{top5}{\makebox[0.9cm][c]{86.26}}   &$\!\!\!\!\!\!$\colorbox{top5}{\makebox[0.6cm][c]{3.37}} &$\!\!\!$\colorbox{top5}{\makebox[0.9cm][c]{183.82}} &\colorbox{top6}{\makebox[0.9cm][c]{854.42}} &$\!\!\!\!\!\!\!\!\!\!\!\!\!$\colorbox{top5}{\makebox[0.9cm][c]{29.99}} &$\!\!\!\!\!\!$\colorbox{top1}{\makebox[0.6cm][c]{2.76}} \\
		LVDM        &$\!\!\!\!\!\!$\colorbox{top5}{\makebox[0.9cm][c]{1271.2}} &$\!\!\!\!$\colorbox{top5}{\makebox[0.9cm][c]{620.41}}  &$\!\!\!\!\!\!\!\!\!\!\!\!\!$\colorbox{top5}{\makebox[0.9cm][c]{97.56}}  &$\!\!\!\!\!\!$\colorbox{top6}{\makebox[0.6cm][c]{1.91}} &$\!\!\!$\colorbox{top6}{\makebox[0.9cm][c]{1007.9}} &$\!$\colorbox{top6}{\makebox[0.9cm][c]{1314.0}} &$\!\!\!\!\!\!\!\!\!\!\!\!\!$\colorbox{top6}{\makebox[0.9cm][c]{109.02}}&$\!\!\!\!\!\!$\colorbox{top6}{\makebox[0.6cm][c]{3.20}} &$\!\!\!$\colorbox{top6}{\makebox[0.9cm][c]{1013.6}} &\colorbox{top5}{\makebox[0.9cm][c]{750.62}} &$\!\!\!\!\!\!\!\!\!\!\!\!\!$\colorbox{top6}{\makebox[0.9cm][c]{196.35}}&$\!\!\!\!\!\!$\colorbox{top6}{\makebox[0.6cm][c]{1.45}}   \\
		MoStGAN-V   &$\!\!\!\!\!\!$\colorbox{top2}{\makebox[0.9cm][c]{437.96}} &$\!\!\!\!$\colorbox{top3}{\makebox[0.9cm][c]{332.93}}  &$\!\!\!\!\!\!\!\!\!\!\!\!\!$\colorbox{top4}{\makebox[0.9cm][c]{53.42}}  &$\!\!\!\!\!\!$\colorbox{top4}{\makebox[0.6cm][c]{3.31}} &$\!\!\!$\colorbox{top3}{\makebox[0.9cm][c]{446.17}} &$\!$\colorbox{top3}{\makebox[0.9cm][c]{461.18}} &$\!\!\!\!\!\!\!\!\!\!\!\!\!$\colorbox{top4}{\makebox[0.9cm][c]{72.61}}   &$\!\!\!\!\!\!$\colorbox{top4}{\makebox[0.6cm][c]{3.56}} &$\!\!\!$\colorbox{top2}{\makebox[0.9cm][c]{89.74}}      &\colorbox{top3}{\makebox[0.9cm][c]{89.11}}  &$\!\!\!\!\!\!\!\!\!\!\!\!\!$\colorbox{top3}{\makebox[0.9cm][c]{15.52}} &$\!\!\!\!\!\!$\colorbox{top2}{\makebox[0.6cm][c]{2.53}}  \\
		Endora      &$\!\!\!\!\!\!$\colorbox{top4}{\makebox[0.9cm][c]{468.37}} &$\!\!\!\!$\colorbox{top4}{\makebox[0.9cm][c]{393.19}}  &$\!\!\!\!\!\!\!\!\!\!\!\!\!$\colorbox{top2}{\makebox[0.9cm][c]{13.75}}  &$\!\!\!\!\!\!$\colorbox{top3}{\makebox[0.6cm][c]{3.86}} &$\!\!\!$\colorbox{top1}{\makebox[0.9cm][c]{383.88}} &$\!$\colorbox{top4}{\makebox[0.9cm][c]{575.53}} &$\!\!\!\!\!\!\!\!\!\!\!\!\!$\colorbox{top3}{\makebox[0.9cm][c]{30.02}}   &$\!\!\!\!\!\!$\colorbox{top3}{\makebox[0.6cm][c]{3.58}} &$\!\!\!$\colorbox{top4}{\makebox[0.9cm][c]{114.36}} &\colorbox{top4}{\makebox[0.9cm][c]{96.68}}  &$\!\!\!\!\!\!\!\!\!\!\!\!\!$\colorbox{top4}{\makebox[0.9cm][c]{18.33}} &$\!\!\!\!\!\!$\colorbox{top5}{\makebox[0.6cm][c]{2.31}}  \\     \midrule
		MedSora-B   &$\!\!\!\!\!\!$\colorbox{top3}{\makebox[0.9cm][c]{462.04}} &$\!\!\!\!$\colorbox{top2}{\makebox[0.9cm][c]{332.23}}  &$\!\!\!\!\!\!\!\!\!\!\!\!\!$\colorbox{top3}{\makebox[0.9cm][c]{15.47}}       &$\!\!\!\!\!\!$\colorbox{top2}{\makebox[0.6cm][c]{3.88}} &$\!\!\!$\colorbox{top4}{\makebox[0.9cm][c]{447.78}} &$\!$\colorbox{top2}{\makebox[0.9cm][c]{447.69}} &$\!\!\!\!\!\!\!\!\!\!\!\!\!$\colorbox{top2}{\makebox[0.9cm][c]{24.98}}   &$\!\!\!\!\!\!$\colorbox{top2}{\makebox[0.6cm][c]{3.63}} &$\!\!\!$\colorbox{top3}{\makebox[0.9cm][c]{92.36}}           &\colorbox{top2}{\makebox[0.9cm][c]{86.43}}       &$\!\!\!\!\!\!\!\!\!\!\!\!\!$\colorbox{top2}{\makebox[0.9cm][c]{12.06}}                            & $\!\!\!\!\!\!$\colorbox{top4}{\makebox[0.6cm][c]{2.38}}               \\
		MedSora-L   &$\!\!\!\!\!\!$\colorbox{top1}{\makebox[0.9cm][c]{429.35}}       &$\!\!\!\!$\colorbox{top1}{\makebox[0.9cm][c]{310.58}}        &$\!\!\!\!\!\!\!\!\!\!\!\!\!$\colorbox{top1}{\makebox[0.9cm][c]{13.42}}       &$\!\!\!\!\!\!$\colorbox{top1}{\makebox[0.6cm][c]{3.93}}    &$\!\!\!$\colorbox{top2}{\makebox[0.9cm][c]{400.60}}       &$\!$\colorbox{top1}{\makebox[0.9cm][c]{412.52}}       &$\!\!\!\!\!\!\!\!\!\!\!\!\!$\colorbox{top1}{\makebox[0.9cm][c]{20.02}}        &$\!\!\!\!\!\!$\colorbox{top1}{\makebox[0.6cm][c]{3.98}}     &$\!\!\!$\colorbox{top1}{\makebox[0.9cm][c]{76.71}}           &\colorbox{top1}{\makebox[0.9cm][c]{80.50}}       &$\!\!\!\!\!\!\!\!\!\!\!\!\!$\colorbox{top1}{\makebox[0.9cm][c]{10.48}}                            &   $\!\!\!\!\!\!$\colorbox{top3}{\makebox[0.6cm][c]{2.44}}             \\
		\bottomrule
	\end{tabularx}
\end{table}

\subsection{Qualitative Evaluation}
\begin{figure}[ht]
	\centering
	\includegraphics[width=\linewidth]{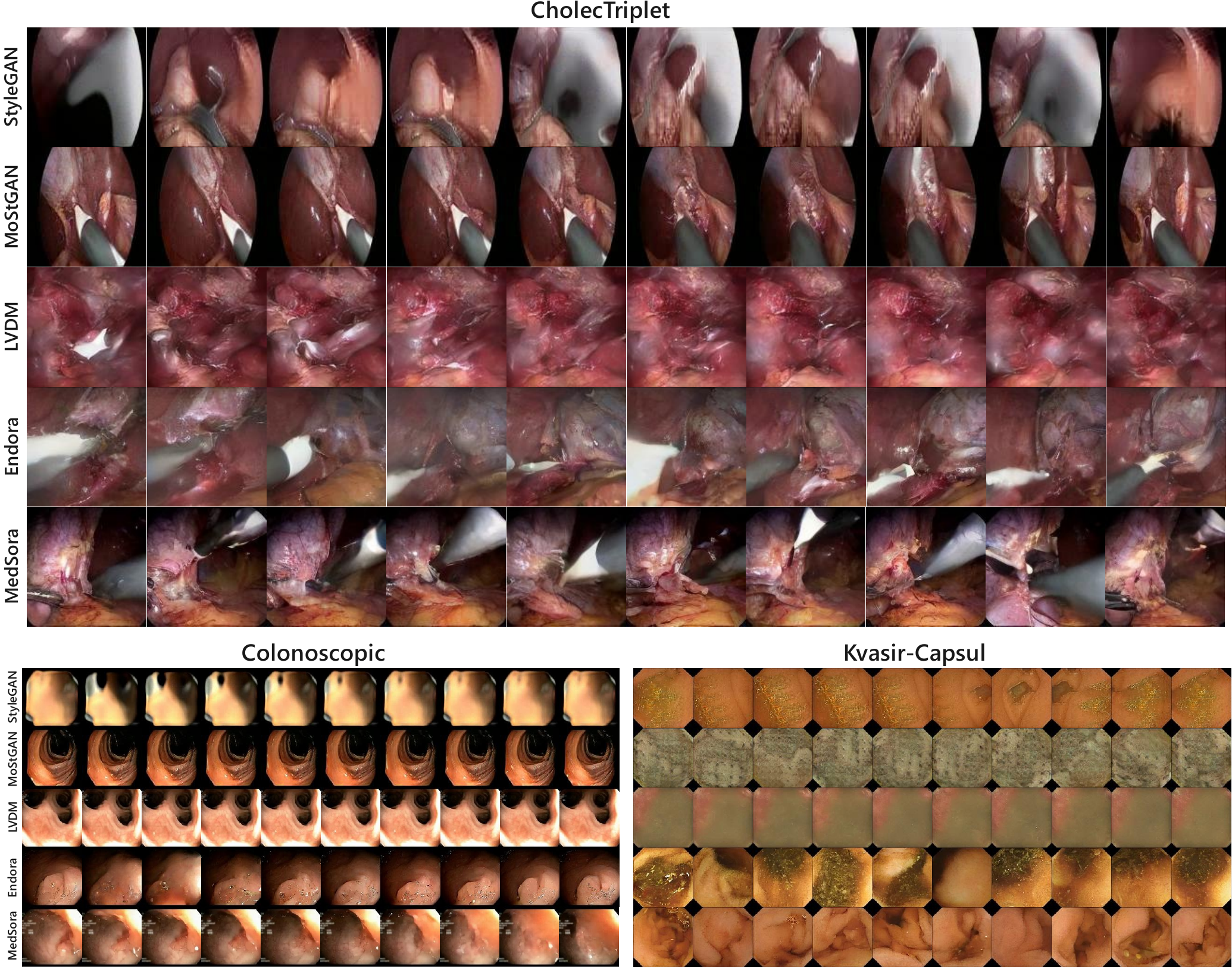}
	\caption{Qualitative comparison between \modell and other video generation models. The generated videos consist of $16$ frames, from which $10$ consecutive frames starting at $t=0$ are selected for demonstration. For a more comprehensive comparison of qualitative results, please refer to the generated videos on our project webpage.}
	\label{fig:exp_compare}
\end{figure}

Figure~\ref{fig:exp_compare} further demonstrates the qualitative results of \modell alongside comparison methods. We observed that other video generation techniques exhibit the following issues: \textit{i)} video distortion (StyleGAN), \textit{ii)} overly consistent frames with insufficient dynamic representation (MoStGAN and LVDM), \textit{iii)} abnormal video imagery that conflicts with anatomical structures (LVDM), and \textit{iv)} video frames display artifacts or blurring, the transitions between frames lack coherence (Endora). 
In contrast, the video frames generated by \modell avoid inconsistent visual distortions, and the transitions between frames are coherent, exhibiting high visual quality and dynamic expression. Furthermore, \modell maintains color consistency and detail clarity, rendering the generated video more visually realistic.

To demonstrate that videos generated by \modell exhibit greater stability and smoother frame transitions than Endora, we computed the FC scores between consecutive frames of videos generated by both models. As presented in Table~\ref{tab:4}, the results indicate that \modell consistently achieves higher FC scores than Endora across the three datasets, confirming its superior frame consistency.

\begin{table}[htbp] \centering
	\caption{Comparison of FC scores between \modell and Endora.} \vspace{0.5em}
	\label{tab:4}
	\begin{tabular}{@{}lccc@{}}
		\toprule
		Method          & Colonoscopic   & CholeTriplet & Kvasir-Capsul  \\ \midrule
		Endora          & 95.43          & 91.15        & 90.70          \\ 
		MedSora-B       & 97.17          & 94.19        & 94.02          \\ \bottomrule
	\end{tabular}
\end{table}

\subsection{Training Efficiency}
\begin{figure}[ht]
	\centering
	\includegraphics[width=\linewidth]{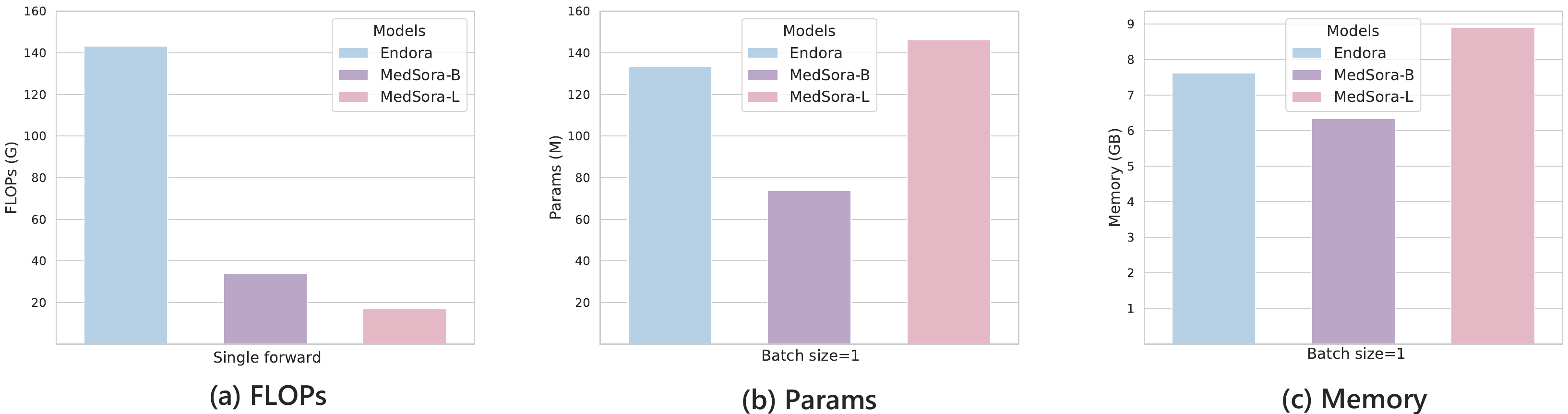}
	\caption{Comparison of training efficiency. (a) FLOPs, (b) model parameter scale, and (c) memory (GB) of different methods that are trained on $16$ frame videos with resolution of $128 \times 128$ and batch size of $1$. To ensure a fair comparison, we do not apply gradient checkpointing to any of the models.}
	\label{fig:flops}
\end{figure}

Figure~\ref{fig:flops} summarizes the floating-point operations (FLOPs), parameter size, and memory consumption involved in training the video Mamba diffusion model, 
and compare with Endora (base size), a method tailored for medical video generation tasks.
As shown in Figure~\ref{fig:flops}, \model\textit{-B} outperforms Endora even though it uses fewer parameters and less memory. This superior performance is due to our proposed optical flow representation alignment supervision and frequency compensation video VAE, which effectively capture the spatio-temporal information of the video.
In addition, we observed that although \model\textit{-L} has slightly more parameters than Endora, its FLOPs are significantly lower. This is because the use of frequency compensation video VAE enables a higher compression rate than the previous image VAEs, and the carefully designed spatio-temporal Mamba structure avoids the quadratic computational complexity introduced by global self-attention.

\subsection{An Example of Downstream Task}
We investigated the utilization of video generation models in downstream semi-supervised tasks. Specifically, within the PolyDiag~\citep{tian2022contrastive} training set, we randomly selected $40$ videos as labeled data and generated 200 additional videos as unlabeled data to train a disease classification model.
Table~\ref{tab:3} presents the F1 scores of the classification model on the Colonoscopic and CholecTriplet datasets. The results indicate that employing generated videos for semi-supervised training substantially enhances the performance.
Furthermore, In addition, compared to fake videos produced by other video generation methods, those generated by \modell demonstrated superior downstream performance. The result indicates that the videos generated by \modell are more realistic, thereby validating \model's effectiveness as a medical video simulator.

\begin{table}[htbp] \centering
	\caption{Semi-supervised classification F1 score on PolyDiag.} \vspace{0.5em}
	\label{tab:3}
	\begin{tabular}{@{}lcc@{}}
		\toprule
		Method          & Colonoscopic   & CholeTriplet  \\ \midrule
		Supervised-only & 74.26          & 74.70         \\
		LVDM            & 75.92 (+1.66)$~~$  & 77.88 (+3.18) \\
		Endora          & 86.73 (+12.47) & 81.89 (+7.19) \\ \midrule
		MedSora-B       & 88.12 (+13.86) & 83.01 (+8.31) \\ \bottomrule
	\end{tabular}
\end{table}

\section{Conclusion}
This paper introduces \modell, an exploratory framework for medical video generation. Specifically, by considering the advantages and disadvantages of attention and Mamba, we propose a video Mamba diffusion model framework that efficiently models the spatio-temporal information in medical videos. To accurately capture inter-frame pixel motion, we propose optical flow representation alignment to assist in the training of the diffusion model. Additionally, we enhance the 3D causal VAE for efficient compression of medical videos.
In benchmark experiments on medical video datasets, \modell demonstrated excellent performance across various metrics, computational efficiency, and impressive potential as a medical video simulator for downstream tasks.
We anticipate that this work will inspire further research in medical video generation and contribute significant breakthroughs to generative AI in medicine.

\bibliography{conference}
\bibliographystyle{conference}


\end{document}